\documentclass[10pt,a4paper]{article}
\usepackage{arxiv}
\usepackage[utf8]{inputenc}
\usepackage{amssymb}
\usepackage{amsmath}
\usepackage{bm}
\usepackage{mathtools}
\usepackage{url}

\usepackage{multirow}
\usepackage{float}
\usepackage{booktabs}
\usepackage{subcaption,graphicx}

\usepackage{epstopdf}

\usepackage{footmisc}

\usepackage[sort&compress,round,authoryear]{natbib}
\usepackage{color, soul}
\sethlcolor{yellow}

\title{Deep learning for Chemometric and non-translational data}

\author{
  Jacob S\o gaard Larsen\thanks{Corresponding author}, Line Clemmensen\\
  Department of Applied Mathematics and Computer Science\\
  Technical University of Denmark\\
  Richard Petersens Plads, 2800, Lyngby, Denmark\\
  \texttt{\{jasla,lkhc\}@dtu.dk}
}

\begin{document}

\maketitle
\begin{abstract}
    We propose a novel method to train deep convolutional neural networks which learn from multiple data sets of varying input sizes through weight sharing. This is an advantage in chemometrics where individual measurements represent exact chemical compounds and thus signals cannot be translated or resized without disturbing their interpretation. Our approach show superior performance compared to transfer learning when a medium sized and a small data set are trained together. While we observe a small improvement compared to individual training when two medium sized data sets are trained together, in particular through a reduction in the variance.
    \end{abstract}
\section{Introduction}
Spectral data consist of spectroscopic measurements which contain chemical information about the composition of the sample. Spectral data are in large based on underlying continuous processes, much as is the case for image data, and thus we expect deep learning to work well for applications on spectral data such as e.g. Near Infra Red (NIR) data. Deep learning has thus also successfully been applied to spectroscopic data within a variety of fields. E.g. \cite{Risum2019} used a deep convolutional neural network (CNN) to detect different types of peaks in gas-chromatography. \cite{Liu2017} used a deep CNN to classify different chemical substances based on Raman spectra. Transfer learning for spectroscopic data has been attempted in specialized cases. \cite{Liu2018} applied transfer learning to hyperspectral data of soil. They used data from spectral libraries (the data were acquired under laboratory conditions) to pre-train a model, which was then transferred to field data. \cite{Padarian2019} used transfer learning to convert a global soil clay model to a locally calibrated model. Common for both \cite{Liu2018} and \cite{Padarian2019} is that the original and new data sets hold the same wavelengths at the same positions. This is contrary to e.g. image data, where two images can hold the same scene, and thereby the same label, but still have e.g. different zoom levels, translations or rotations.

Often, chemometricians work with data of few samples and large amounts of input variables, which may be one of the reasons deep learning has not gained a broad use in the field yet. A well known strategy to easily gain more samples is by data augmentation, where different artefacts are added to each sample. Typical techniques for image data apply rotation and translation. For spectroscopy data, \cite{Bjerrum2017} proposed to add different types of scattering to the spectra. This strategy was used to train a deep neural net on the 2002 IDRC Challenge Data \citep{Norris2008,Hopkins2003}. However, data augmentation cannot fully compensate for the lack of original training samples, as it only enables the neural net to compensate for the artefacts one add to the original samples. 

Another strategy is to merge multiple data sets. \cite{Ma2015} showed that by merging multiple small QSAR data sets, they were able to learn a much better Deep Neural Net (DNN) with multiple tasks than if the same DNN was trained on the individual data sets. They also found that the gain of merging multiple data sets is larger for many small data sets than e.g. two large data sets.

A third strategy is to use transfer learning, which deep learning has proven useful at due to its focus on learning representations of data \mbox{\citep{Bengio2012}}. Applications using images, text and speech have particularly benefited from the use of deep learning and transfer learning \mbox{\citep{LeCun2015}}. Well known, pre-trained networks such as AlexNet \mbox{\citep{Krizhevsky2012}}, GoogleNet \mbox{\citep{Szegedy2015}}, etc. are available as off the shelf solutions to quickly and without large amounts of data get started on your own image based deep learning applications.

The lack of consistency in the number of wavelengths used when spectral data sets are constructed makes it difficult to merge these data sets into one. Common practice within image analysis is to resize the images. However, due to the continuous nature of the spectra, this is likely to introduce noise, and thereby reduce the final performance of the model. We therefore propose a new strategy, that makes it possible to merge the information from multiple data sets without resizing them. This is done by assigning a deep neural net to each data set, with the restriction that the weights in the convolutional layers are shared among the nets. We note that this strategy has similarities to transfer learning. However, our strategy differ in one key aspect. In transfer learning, the learned representation is transferred from one data set to another. In our approach, all data sets contribute to learning the optimal representation.

This paper is structured as follows: In Section \ref{sec:method} we describe the proposed method for co-training a deep CNN on multiple data sets with different input sizes. In Section \ref{sec:data} we describe the data sets used in this study. Our experimental setup is described in Section \ref{sec:setup}, the results are presented and discussed in Section \ref{sec:results} and Section \ref{sec:conclusion} concludes the paper.

All analyses have been performed in Python version 3.6 (Python Software Foundation, \url{https://www.python.org/}), and Neural Nets have been implemented using TensorFlow version 1.12 \citep{GoogleResearch2015}. Examples of implementing similar Neural Nets using TensorFlow are available at the GitHub repository \\ \url{https://github.com/DTUComputeStatisticsAndDataAnalysis}.
\section{Method} \label{sec:method}
In this section we describe our proposed method for training deep convolutional neural nets on multiple data sets with varying input sizes through Weight Sharing. Subsequently, we present a regularization cost for achieving sparse and decoupled weights.

\subsection{Weight Sharing}
Consider the 1d convolution between a signal $\bm{x}$ of length $p$ and a filter of length $2h+1$ given by the parameters $\bm{\theta}$ in Eq. \eqref{eq:1d_conv} (with a suitable padding strategy at the endpoints). Realizing that the length of the input signal is not the limiting factor, a convolution could also be performed on another signal $\bm{x}^*$ of length $p^*$ using the same filter.

\begin{equation} \label{eq:1d_conv}
    \left(\bm{x} * \bm{\theta} \right)_i = \sum_{j=0}^{2h} x_{i+h-j} \theta_{j+1},\qquad i\in [1,p]
\end{equation}

\begin{figure}[ht]
    \centering
    \includegraphics[width=\textwidth]{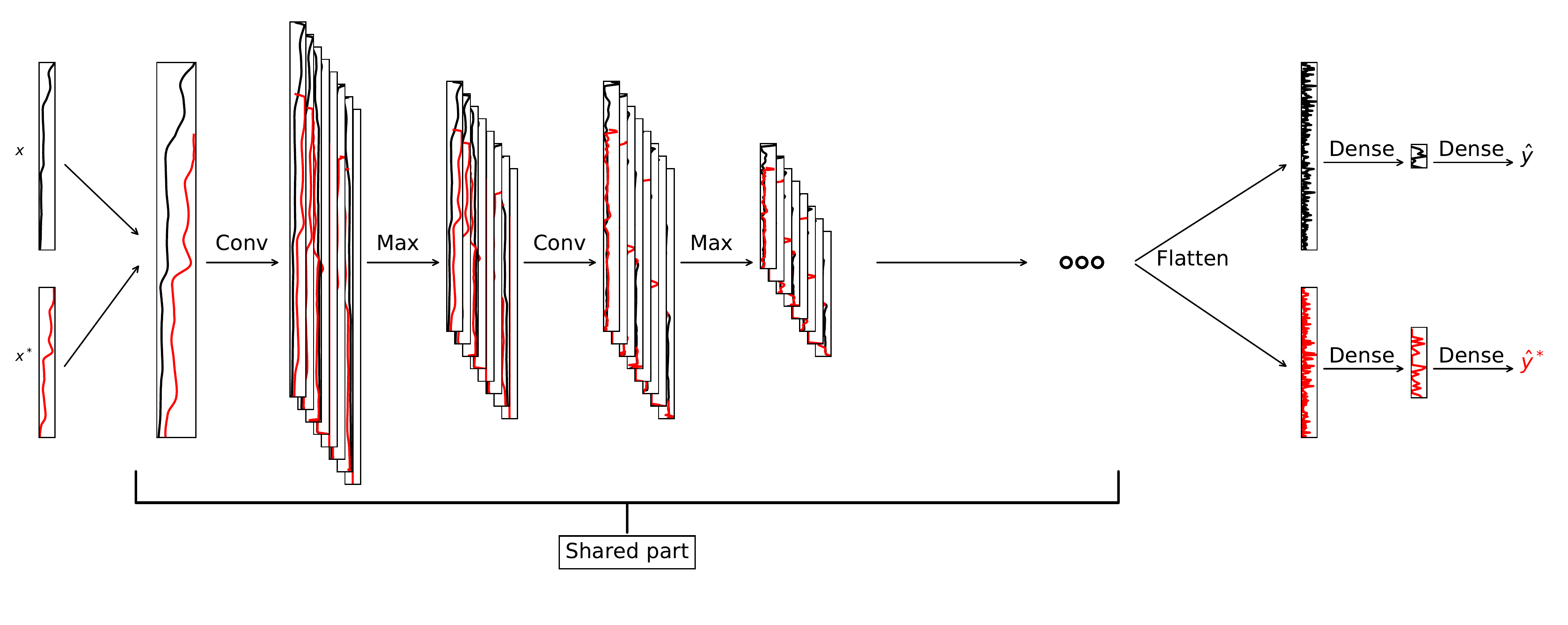}
    \caption{Illustration of the proposed architecture and how two spectra $x$ and $x^*$ of different input sizes is passed through the network.}
    \label{fig:CNN_arch}
\end{figure}

\noindent Based on this, we propose to create multiple neural nets with the same overall architecture, but with varying numbers of input variables, where the weights of the convolutional layers are shared. In this way, one can learn higher level representations that generalize to multiple data sets regardless of the data sets having different input sizes. The strategy is illustrated in Fig. \ref{fig:CNN_arch}, where samples of different input sizes pass through the same convolutional layers. After the convolutional layers, the net is split into two, with separate fully connected layers.

When training the nets, the $k$'th net has its own cost function $d_k(\mathbf{X}^k,\mathbf{y}^k;\bm{\Theta}^k)$, with $(\mathbf{X}^k,\mathbf{y}^k)$ being the $k$'th data set and $\bm{\Theta}^k$ being the parameters related to the $k$'th net. One could combine the cost functions into one by the weighted sum $d((\mathbf{X}^1,\mathbf{y}^1),(\mathbf{X}^2,\mathbf{y}^2),\dots;\bm{\Theta}^1, \bm{\Theta}^2,\dots) = \sum_{k=1}^K \alpha_i d_k(\mathbf{X}^k,\mathbf{y}^k;\bm{\Theta}^k)$. However, this would require tuning of the weight parameters $\alpha_i$, as a non-optimal choice would result in one cost function dominating the sum. Instead, we propose to use an alternating approach, where we alternate between updating each individual net (and shared weights). Besides avoiding the need to tune $\alpha$, in our experience, this also helps in avoiding local minima.

\subsection{Sparse and decoupled weights}
In the case of multi-task learning, it can be of interest to enforce independence of the weight vectors for each output if the individual outputs are independent. Let the $k$'th layer have $p^{(k)}$ units, the $(k+1)$'th layer have $p^{(k+1)}$ units and $\bm{\theta}^{(k)} = [\theta^{1,k},\theta^{2,k},\dots]$ be the $p^{(k)}\times p^{(k+1)}$ weight matrix associated with with the $k$'th layer. For each row of $\bm{\theta}^{(k)}$ we only want one non-zero value. This can be achieved by using the regularization cost given in Eq. \eqref{eq:regu_independence}, where $\lambda\geq 0$ is the regularization parameter. In our experience the main issue regarding tuning $\lambda$ is that it is not set too high, such that the regularization cost dominates the cost function in the beginning of training, which limits the initial learning by forcing the parameters towards zero.

\begin{equation} \label{eq:regu_independence}
    \Omega \left(\bm{\theta}^{(k)};\lambda \right) = \lambda \sum_{i=1}^{p^{(k+1)}-1} \sum_{i'=i}^{ p^{(k+1)} } \sum_{l=1}^{p^{(k)}} |\theta^{i,k}_l \theta^{i',k}_l|
\end{equation}

\section{Data} \label{sec:data}
This section describes the data sets used in this paper. They consist of NIR measurements from silage, a mixture of food substances (soya oil, lucerne and barley), pharmaceutical tablets, wheat kernels and diesel fuels, respectively.  All data sets are measured at different numbers of wavelengths, but all in the NIR region of the electromagnetic spectrum.

\subsection{Chimiometrie 2018}
This data set was first published as challenge data at the Chimiometrie 2018 conference in Paris and is available at the conference homepage\footnote{\url{https://chemom2018.sciencesconf.org/resource/page/id/5}}. The data set consists of NIR measurements from 10 different types of silage measured at 680 (unknown) wavelengths. The target was to predict the protein content, as the silage type was not provided for each measurement. A calibration set consisting of 3908 unique spectra and corresponding target values were provided. Furthermore, 429 test spectra were provided without the target values. Out of the 429 test spectra, 57 spectra were also present in the training data. However, the participants were not informed of this.

\subsubsection{Benchmarks}
At the challenge, the winning calibration was judged by its median absolute deviance (MAD). However, as this is invariant to an additive constant, we will also use the root mean squared error (RMSE) of prediction to evaluate the the models.

In Table \ref{tab:benchChim2018}, the MAD and RMSE of both the winner of the challenge (Winner) and the data providers' own solution (CRA-W) are shown. The winner of the challenge has the best performance and used Gaussian Process Regression and pre-processed the data using a Standard Normal Variate (SNV) transformation on top of a first order Savitzky-Golay derivative.

\begin{table}[htb]
    \centering
    \begin{tabular}{ccc}
        \toprule
        &  RMSE & MAD\\
        \midrule
        CRA-W & 0.69 & 0.385 \\
        Winner & 0.687 & 0.365 \\
        \bottomrule
    \end{tabular}
    \vspace{.1cm}
    \caption{Benchmarks of the Chimiometrie 2018 data set.}
    \label{tab:benchChim2018}
\end{table}

\subsection{Chimiometrie 2019}
This data set was published as the challenge at the Chimiometrie 2019 conference held in Montpellier and is available at the conference homepage\footnote{\url{https://chemom2019.sciencesconf.org/resource/page/id/13}}. The data consist of 6915 training spectra and 600 test spectra measured at 550 (unknown) wavelengths. The target was the amount of soy oil ($0-5.5\%$), lucerne ($0-40\%$) and barley ($0-52\%$) in a mixture. The test set was measured using a different instrument, resulting in a shift in the test spectra of 0.5nm, making the challenge harder.

\subsubsection{Benchmarks}
At the challenge, the objective was to minimize the Weighted RMSE (WRMSE) given in Eq. \eqref{eq:wrmse_2019}, with $\bar{y}_{soya\ oil}$, $\bar{y}_{lucerne}$ and $\bar{y}_{barley}$ being the average amount of soy oil, lucerne and barley in the training samples.

\begin{equation} \label{eq:wrmse_2019}
WRMSE = \frac{1}{3} \left( \frac{RMSE_{soya\ oil}}{\bar{y}_{soy\ oil}} + \frac{RMSE_{lucerne}}{\bar{y}_{lucerne}}+\frac{RMSE_{barley}}{\bar{y}_{barley}}\right)
\end{equation}

In Table \ref{tab:benchChim2019}, the WRMSE of both the data providers (UCO\footnote{University of Cordoba, Prof. Ana Garrido}) and the winner of the challenge (Winner) are shown. The data provider has the best performance and used a combination of Standard Normal Variate (SNV) and 1st order Savitzky Golay filtering as pre-processing of the data. They then used LOCAL \citep{Shenk1997} as the calibration method.

\begin{table}[htb]
    \centering
    \begin{tabular}{cc}
        \toprule
        &  WRMSE\\
        \midrule
        UCO & 0.64 \\
        Winner & 0.70 \\
        \bottomrule
    \end{tabular}
    \vspace{.1cm}
    \caption{Benchmarks of the Chimiometrie 2019 data set.}
    \label{tab:benchChim2019}
\end{table}

\subsection{IDRC 2002}
This data set was first introduced at the 2002 International Diffuse Reflectance Conference (IDRC) as a challenge. However, as the homepage is no longer available, we collected the data at Eigenvector's homepage\footnote{\url{https://eigenvector.com/resources/data-sets/}\label{eigenvector_data}}. The data consist of NIR measurements of a total of 655 pharmaceutical tablets each measured on two different instruments at 650 wavelengths with the objective to predict the amount of active ingredient (API). For each tablet the weight of both the tablet and the total amount of API are provided. The data set is divided into a training set consisting of 155 tablets, a validation set of 460 tablets and a test set of 40 samples. Further details of the data are described in \cite{Hopkins2003}.  We note that in this study, we only use data from instrument 1. Furthermore, we use the 460 validation samples as our test set, and combine the original training set and test set into one training set of 195 samples.

\subsection{Wheat}
This data set was published together with the papers \cite{Pedersen2002,Nielsen2003} and is available at the KU FOOD Quality and Technology homepage\footnote{\url{http://models.life.ku.dk/wheat_kernels}}. The data consist of NIR spectra samples of wheat kernels collected from different locations measured at 100 wavelengths. The data set is divided into a training set of 415 samples and a test set of 108 samples.

\subsubsection{Benchmarks}
This data set has been used in several studies. In Table \ref{tab:benchWheat} the RMSE of both the data providers and the current benchmarks for linear and non-linear methods are shown. It is clear that not much is gained from changing into a non-linear method like CNN, we therefore suspect that the signal of interest is linear.

\begin{table}[htb]
    \centering
    \begin{tabular}{lrr}
        \toprule
        & Method & RMSE \\
        \midrule
        \citeauthor{Pedersen2002,Nielsen2003} & PLS & 0.48 \\
        \citeauthor{Cui2018} & PLS & 0.425 \\
        \citeauthor{Cui2018} & CNN & 0.420 \\
        \bottomrule
    \end{tabular}
    \vspace{.1cm}
    \caption{Benchmarks for the Wheat data set.}
    \label{tab:benchWheat}
\end{table}

 \subsection{SWRI}
 This data set was built by the Southwest Research Institute (SWRI) in order to evaluate fuel on the battle fields, however we collected the data at Eigenvectors homepage\footref{eigenvector_data}. The data set consist of 784 raw spectra of different diesel fuels. For each sample several properties have been measures such as boiling point, total aromatics mass in \% etc. However, not all properties has been measured for all samples, i.e. there are a lot of missing values. We have chosen to predict the total aromatics mass in \%, for which there are 395 samples. The data set does not come with a dedicated test set on the raw spectra.

\section{Experimental Setup} \label{sec:setup}
In this section we describe the setup of the experiments conducted in this study, the training strategy used and how we will evaluate and compare the final performance. For all studies we consider two architectures of the neural nets, where the difference lies in the filter length of the shared part. Furthermore, we add two fully connected layers separated by a batch normalization layer on top of the shared part. The parameters of the fully connected layers are not shared among the nets. For details on the architecture see Tables \ref{tab:Common_arch_1} and \ref{tab:fc_part} in Appendix \ref{app:Arch}.

\subsection{Experiment 1: Weight sharing for two medium sized data sets} \label{sec:weight_share}
We train on the Chimiometrie 2018 and 2019 data sets with shared weights among the convolutional layers. The nets are updated $50,000$ times with a batch size of $128$ samples from each data set and an initial learning rate of $10^{-3}$, which is dropped by a factor of 2 when there hasn't been an improvement in the validation error for 10 epochs - this is done until a minimum learning rate of $3\cdot 10^{-5}$ is reached. 

To asses the performance of Weight Sharing, as baseline we perform the same experiments with individual training instead of co-training.

The performance is evaluated using RMSE and MAD on the Chimiometrie 2018 data set and WRMSE and biases of the three targets on the Chimiometrie 2019 data set.

\subsection{Experiment 2: Weight sharing for a small and a medium sized data set}
We train on the Chimiometrie 2019 data set and a smaller data set and share the weights among the convolutional layers. We use the same training strategy as outlined in Section \ref{sec:weight_share}.

Besides being used in a co-training setting, the proposed method can also be used for transfer learning, even though the pre-trained net doesn't have the same number of input variables as the smaller data set. We do this by picking the two best performing nets from the medium sized data sets in Section \ref{sec:weight_share} trained individually. We then transfer the parameters of the convolutional layers and subsequently train the network with the smaller data set. We employ two strategies for updating the parameters. 1. TL WS Stop Gradient: We only update the fully connected layers with the smaller data set. 2. TL WS Full Gradient: We update the entire net using the smaller data set. We update the nets for 200 epochs using a batch size of 128 and an initial learning rate of $10^{-3}$, which is dropped by a factor of 2 when there hasn't been any improvement for 50 epochs, until a minimum learning rate of $3\cdot 10^{-5}$ is been reached.

As a baseline, we perform traditional transfer learning, where we either pad the spectra on both sides or interpolate using cubic splines, such that the input size matches that of the pre-trained net. We employ the two training strategies as described above, naming them 1. TL Stop Gradient and 2. TL Full Gradient.

We evaluate the performance of the strategies using RMSE, Standard Error of Prediction (SEP), R$^2$ and Bias.

\subsection{Data splits} \label{sec:data_split}
For each experiment, we perform 40 repetitions in order to assess the statistical properties of the strategies. For each repetition of the experiments we sub-sample our training data into three data sets: training data used to train the models, validation data used during training to decide at which iteration to store the model and a hold out data set used after training to select the architecture. Recall that in all cases, except for the SWRI data set, the test set is fixed and used to evaluate the performance of the opposing strategies for each experiment. The number of samples in each data set is given in Table \ref{tab:data_split}. We note that for the SWRI data set the test data are not overlapping between repetitions. Given a repetition number, we use the same sub-sampled data set for each of the strategies, this produce paired experiments, which will be utilized in the analysis of the results.

\begin{table}[htb]
    \centering
    \begin{tabular}{lrrrrr}
        \toprule
        Data set & Input size & Training data & Validation data & Hold out data & Test data \\
        \midrule
        IDRC 2002 & 650 & 140 & 35 & 20 & 460\\
        Wheat & 100 & 298 & 75 & 42 & 108\\
        SWRI & 401 & 276 & 70 & 39 & 10 \\
        Chimiometrie 2018 & 680 & 2813 & 704 & 391 & 429\\
        Chimiometrie 2019 & 550 & 4978 & 1245 & 692 & 600\\
        \bottomrule
    \end{tabular}
    \vspace{.1cm}
    \caption{Input sizes and number of samples in each data split}
    \label{tab:data_split}
\end{table}

Prior to training we scale up the training and validation data by a factor of 10 using the data augmentation strategy described in \citep{Bjerrum2017} and append it to the original data. However, we do not augment the hold out data used to select between the architectures.

\subsection{Optimization strategy}
For all our experiments we use the Adam optimizer \citep{Kingma2014}. During training we keep track of an exponential moving average smoothing of our parameters as shown in Eq. \eqref{eq:ema_parameters} with a decay rate of $\gamma=0.99$. The exponential moving averaged parameters are used to evaluate the validation samples, and we store parameters minimizing the sum of the validation error of each of the neural nets. We use the exponentially smoothed parameters to achieve a more stable estimate of the parameters. We note that the exponentially smoothed model is similar to the Teacher model proposed by \cite{Tarvainen2017}, where the difference is that we do not penalize the difference in the prediction between the student and teacher models. Finally, we choose among the opposing strategies using a held out data set, as described in Section \ref{sec:data_split}. For all the trained neural nets we use the Rectified Linear Unit (ReLU) \citep{Jarrett2009,Nair2010} as activation function.

\begin{equation} \label{eq:ema_parameters}
    \tilde{\Theta}^t = \gamma \tilde{\Theta}^{t-1} + (1-\gamma) \Theta^t
\end{equation}

In many spectroscopic applications, it is common to pre-process the spectra. However, as shown by \cite{Cui2018}, a CNN is able to automatically learn an appropriate pre-processing of the spectra. Therefore, we do not perform any pre-processing.

\subsection{Cost function}
For the IDRC 2002, Wheat, SWRI and Chimiometrie 2018 data sets we use RMSE as the training and validation cost functions. For the Chimiometrie 2019 data set we use the WRMSE added the regularization cost in Eq. \eqref{eq:regu_independence}, with $\lambda = 0.1$, for both training and validation cost functions.

\subsection{Pairwise comparisons}
In Experiment 1 we perform a pairwise comparison of the two training strategies based on MAD, RMSE, WRMSE and Bias. For this we use the Wilcoxon signed-ranks test \citep{Wilcoxon1945}. Note than when comparing biases, we compare the absolute values, with the smaller the better.

\subsection{Multiple comparison of strategies}
In Experiment 2 we are comparing five strategies simultaneously, for this we use the Friedman Test \citep{Friedman1937,Friedman1939} with the improved statistic by \cite{Iman1980}. This is done on the measures RMSE, SEP, $R^2$ and Bias.

As a post hoc analysis we use the Nemenyi test \citep{Nemenyi1963} to measure if two rankings are significantly different.

\section{Results} \label{sec:results}
This section presents the results from the two experiments. Besides the results presented here, summary statistics of the experiments are given in Tables \ref{tab:weight_share_stats}, \ref{tab:idrc_transfer_stats} and \ref{tab:wheat_transfer_stats} in Appendix \ref{app:add_res}.

\subsection{Experiment 1} 
\begin{figure}[p]
    \centering
    \includegraphics[width=1\textwidth]{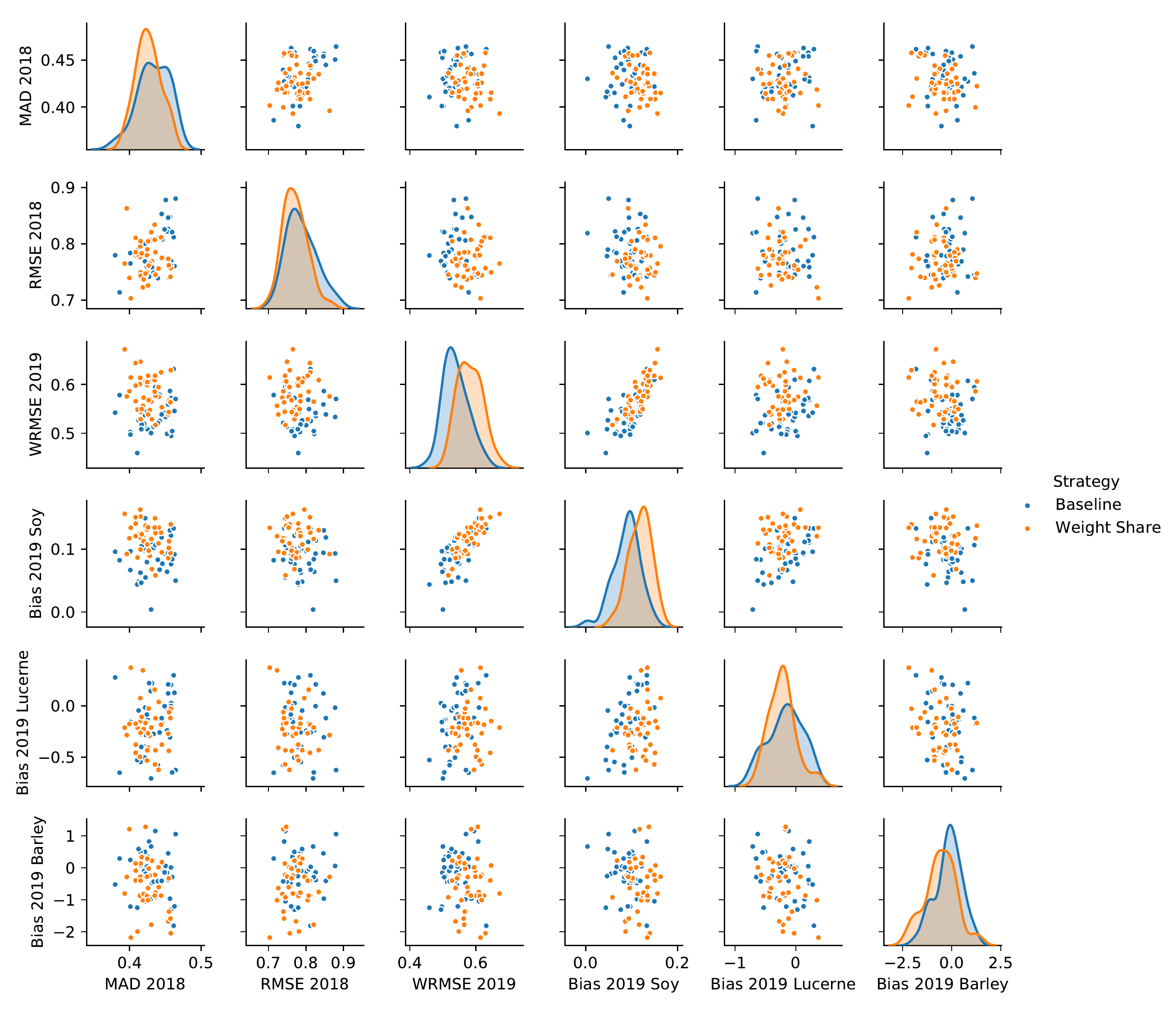}
    \caption{Pairs plot of the performance measures for 40 repetitions of individual training (baseline) and co-training (Weight Share) for the Chimiometrie 2018 and 2019 data sets.}
    \label{fig:weigth_share_vs_baseline}
\end{figure}

Figure \ref{fig:weigth_share_vs_baseline} illustrates a pairs plot and kernel density estimates of the performance metrics for the baseline (individually trained models) and the Weight Sharing strategy for the 40 repetitions. For RMSE 2018 it is seen that the mode is shifted to the left for the weight sharing strategy, i.e. to smaller values, while the opposite is the case for WRMSE 2019. It is clear that the reason is that the baseline models have a smaller bias when predicting the amount soy oil (Bias 2019 Soy). Furthermore, we see that for the baseline models, MAD 2018 has a plateau like mode while the Weight Share model has a sharp peak. Finally, in 37 cases the Weight Sharing strategy had a WRMSE on the Chimiometrie 2019 data set which was smaller than the benchmark, while this was the case for all 40 cases for the baseline. For the Chimiometrie 2018 data set, none of the strategies were able to beat either the MAD or RMSE benchmarks.

We perform a Wilcoxon signed-rank test on MAD and RMSE for the Chimiometrie 2018 data set and WRMSE and the absolute bias for the three targets for the Chimiometrie 2019 data set. The test values and p-values are shown in Table \ref{tab:weight_share_wilcoxon}. It is seen that the Weight Share strategy is significantly better in terms of RMSE on the Chimiometrie 2018 data set, while the baseline is better in terms of WRMSE and bias when predicting the amount of soy oil for the Chimiometrie 2019 data set. The reason for this is that there is a large noise component in the 2019 test data set, as the test data is measured using a different instrument, which caused a shift of 0.5nm of the spectra.

Performing an F-test for a change in variance of all the considered performance metrics produces the test statistics and corresponding p-values presented in Table \ref{tab:weight_share_F_tests}. It is seen that in neither case there is a significant change in variance of the performance metrics between the two strategies.

\begin{table}[p]
         \begin{minipage}[t]{.55\linewidth} 
            \begin{table}[H]
            \centering
                \begin{tabular}{llrr|rr}
                    \toprule
                    {} & Dataset & Test val & p value &    R$_+$ &    R$_-$ \\
                    \midrule
                    MAD & 2018    &   -1.841 &   0.066 &  547 &  273 \\
                    \textbf{RMSE} & 2018   &   -2.621 &   0.009 &  \textbf{605} &  215 \\
                    \textbf{WRMSE} & 2019  &   -4.355 &   0.000 &   86 &  \textbf{734} \\
                    \textbf{Bias Soy oil} & 2019 &   -4.274 &   0.000 &   92 &  \textbf{728} \\
                    Bias Lucerne & 2019 &  -0.470 &   0.638 &  375 &  445 \\
                    Bias Barley & 2019 & -1.532 &   0.125 &  296 &  524 \\
                    \bottomrule
                \end{tabular}
                \vspace{.1cm}
                \caption{\parbox{.75\linewidth}{Test statistic, p-value and effect sizes for the Wilcoxon signed-rank test with significant tests in bold. R$_+$(R$_-$) is the sum of ranks for which the Weight Share(baseline) strategy perform better.}}
                \label{tab:weight_share_wilcoxon}
            \end{table}
         \end{minipage}
         \hfill
         \begin{minipage}[t]{.45\linewidth} 
            \begin{table}[H]
                \centering
                \begin{tabular}{llrr}
                    \toprule
                    Metric  &   Data set &   Test Value  &   p-value  \\
                    \midrule
                    MAD & 2018          &      1.585 &   0.155 \\
                    RMSE & 2018         &      1.457 &   0.245 \\
                    WRMSE & 2019        &      1.027 &   0.934 \\
                    Bias Soy oil & 2019 &      1.464 &   0.238 \\
                    Bias Lucerne & 2019 &      1.631 &   0.131 \\
                    Bias Barley  & 2019  &      0.729 &   0.327 \\
                    \bottomrule
                \end{tabular}
                \vspace{.1cm}
                \caption{\parbox{.75\linewidth}{Test statistics and p-values for F test on change in variance of performance metrics. Significant tests are highlighted with bold faced numbers. The test is performed such that a test value larger than 1 corresponds to a variance reduction from the baseline to the Weight Share strategy.}}
                \label{tab:weight_share_F_tests}
            \end{table} 
         \end{minipage}
\end{table}

\subsection{Experiment 2}
\begin{figure}[H]
    \centering
    \includegraphics[width=.75\textwidth]{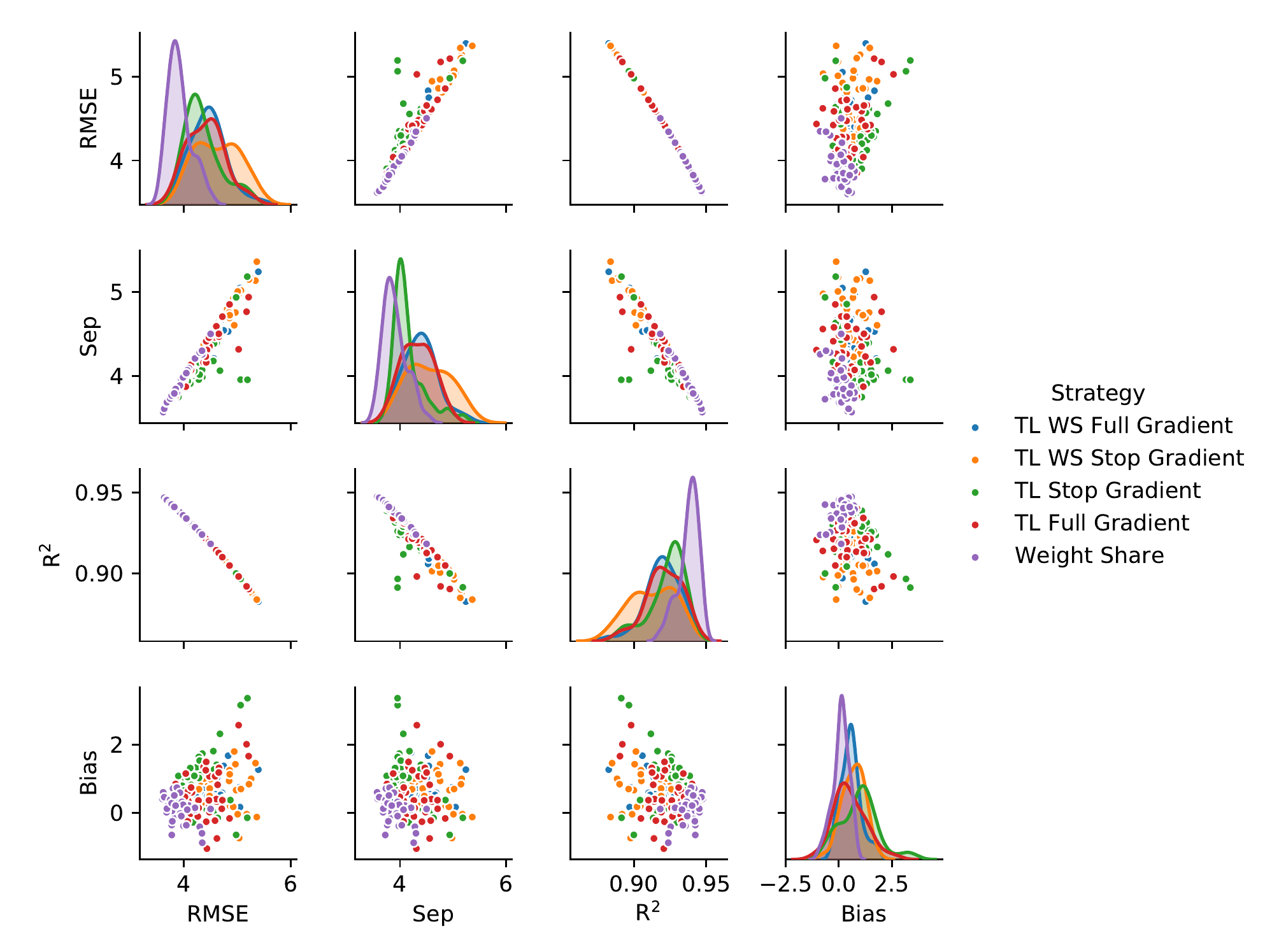}
    \caption{Pairwise plots of the performance metrics for the different transfer learning strategies and the proposed Weight Sharing strategy on the IDRC 2002 data set.}
    \label{fig:transfer_vs_baseline}
\end{figure}

A pairs plot of the performance metrics for the IDRC 2002 data set is shown in Figure \ref{fig:transfer_vs_baseline}. For all metrics the mode for Weight Sharing is better than all four transfer learning strategies. Further, it is clear that the transfer learning strategies produce distributions with long tails of large statistics.

\begin{figure}[H]
    \centering
    \includegraphics[width=.75\textwidth]{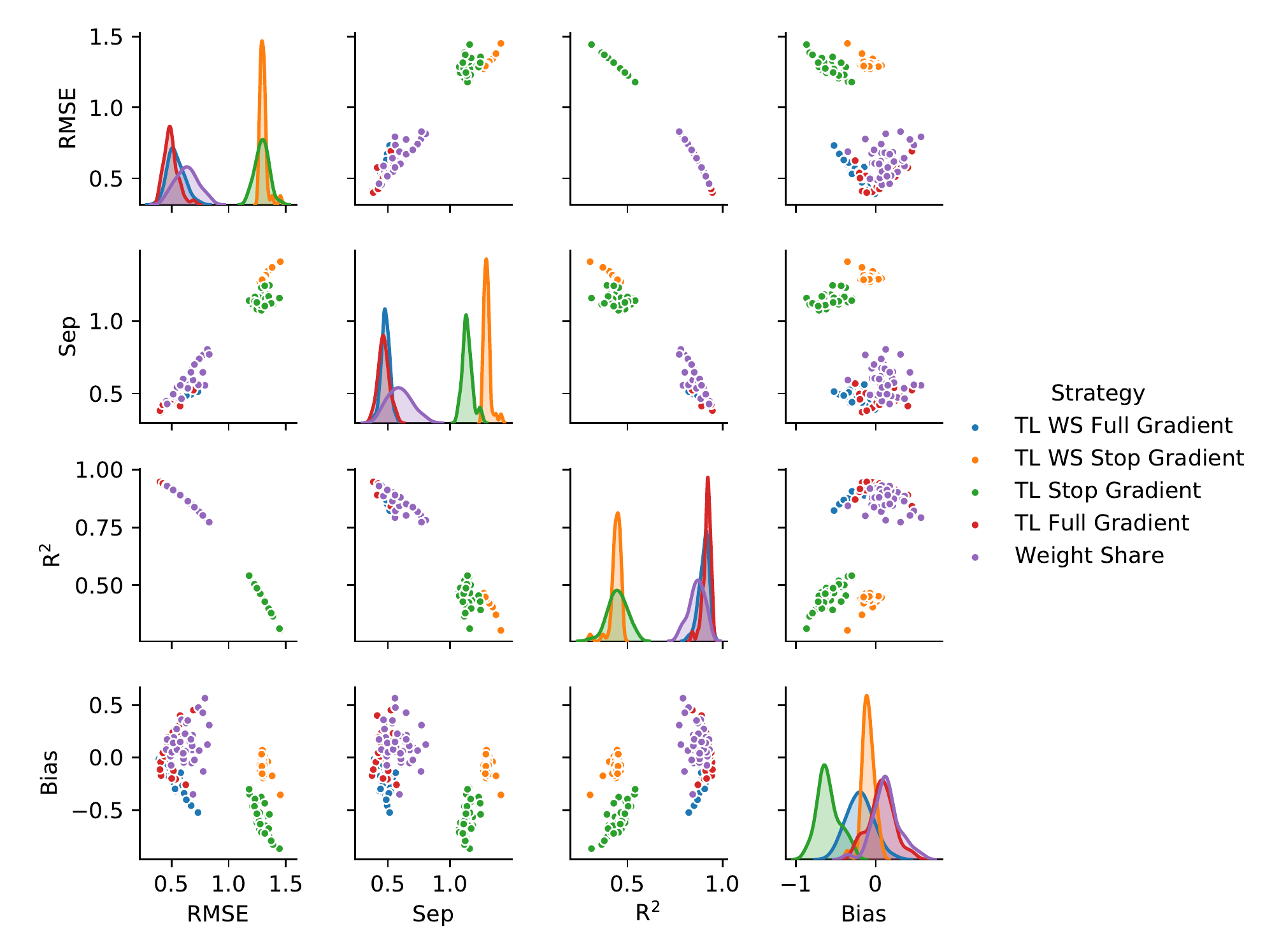}
    \caption{Pairwise plots of the performance metrics for the different transfer learning strategies and the proposed Weight Sharing strategy on the Wheat data set.}
    \label{fig:WHEAT_transfer_vs_baseline}
\end{figure}

Figure \ref{fig:WHEAT_transfer_vs_baseline} illustrates a pairs plot of the performance metrics for the Wheat data set. First it is noted that the both Stop Gradient strategies are clearly separated from the other three strategies. Further, the two Full Gradient strategies are overlapping, with modes sligtly better than that of Weight Share. We also note that for none of the strategies the modes are close to the benchmark. This is expected as we, contrary to the reported benchmarks, are using a subset of the already limited amount of training data to train our models (although in 1 case for TL WS Full Gradient and 4 cases for TL Full Gradient the performance is actually better than the benchmarks).

\begin{figure}[H]
    \centering
    \includegraphics[width=.75\textwidth]{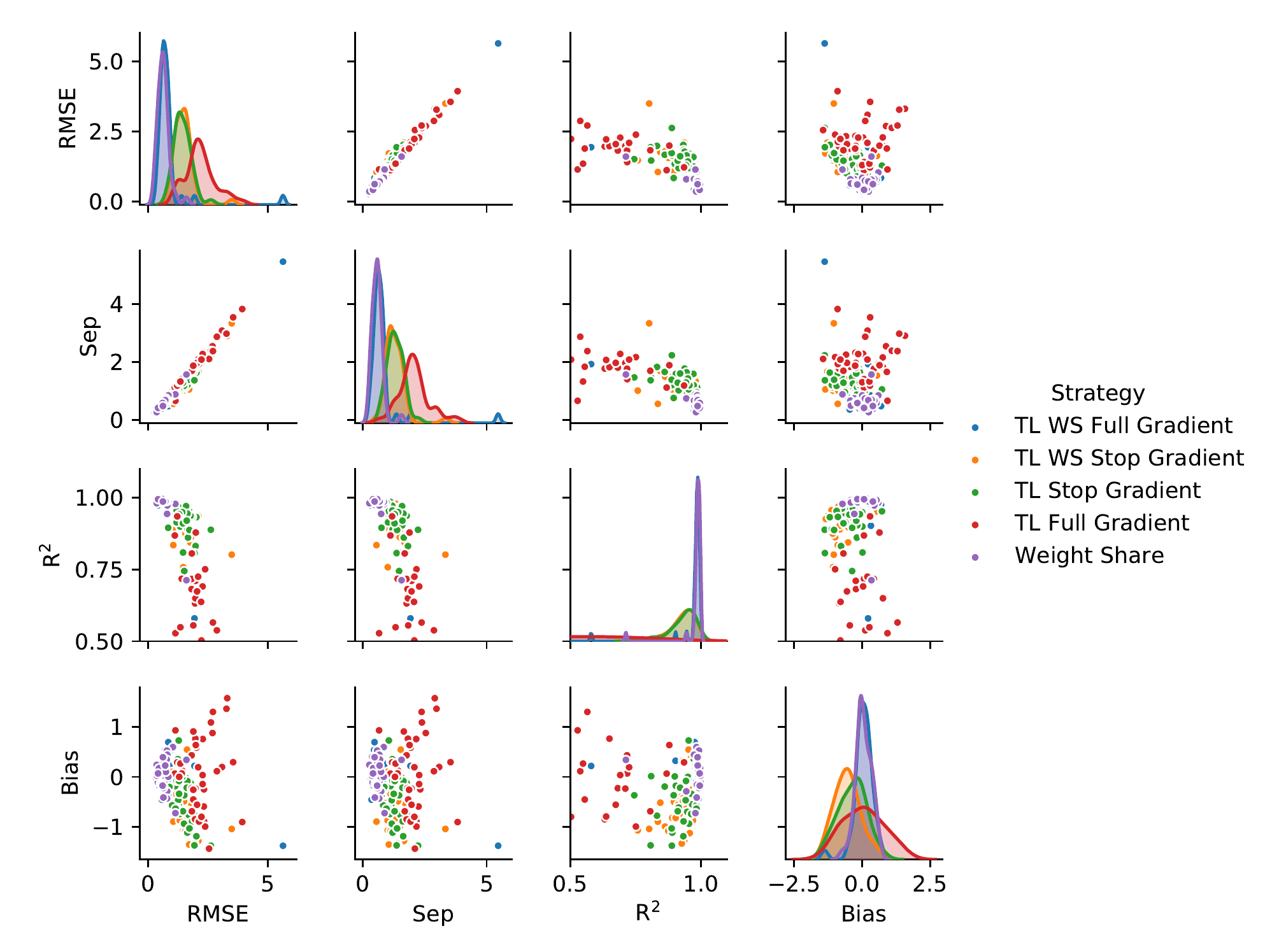}
    \caption{Pairwise plots of the performance metrics for the different transfer learning strategies and the proposed Weight Sharing strategy on the SWRI data set.}
    \label{fig:SWRI_transfer_vs_baseline}
\end{figure}

Figure \ref{fig:SWRI_transfer_vs_baseline} illustrates the performance metrics for the SWRI data set in a pairs plot. It is seen that, for all metrics, the Weight Share and TL WS Full Gradient strategies are overlapping, with the mode of Weight Share being slightly better in all cases. Furthermore, it is seen that the other transfer learning strategies produce distributions with long tails of large statistics.

The test statistics and p-values when testing for a change in variance of the performance metrics is given in Table \ref{tab:transfer_F_tests}. It is clear that for the IDRC and SWRI data sets, a significant decrease of variance has occurred in most cases. For the Wheat data set, an increase of the variance is observed in most cases - 11 out 16 being statistically significant. Here we note that, as seen in Figure \ref{fig:WHEAT_transfer_vs_baseline}, the two Stop Gradient strategies are consequently performing worse than the three other strategies.

\begin{table}[H]
    \centering
    \begin{tabular}{rr|ll|ll|ll}
        \toprule
            {} & {} & \multicolumn{2}{c}{IDRC} & \multicolumn{2}{c}{Wheat} & \multicolumn{2}{c}{SWRI} \\
            Metric  & Baseline              & Test Value & p-value & Test Value & p-value & Test Value & p-value \\
            \midrule
            RMSE    & TL WS Stop Grad.            &      \textbf{3.384} &   \textbf{0.000} &       \textbf{0.116} &   \textbf{0.000} &     \textbf{ 3.057} &   \textbf{0.001} \\
                    & TL WS Full Grad.            &      \textbf{2.215} &   \textbf{0.015} &       			0.530 &   0.051  &     \textbf{11.245} &   \textbf{0.000} \\
                    & TL Stop Grad.     &      \textbf{2.379} &   \textbf{0.008} &       \textbf{0.351} &   \textbf{0.002} &     \textbf{ 2.060} &   \textbf{0.026} \\
                    & TL Full Grad.     &      \textbf{2.397} &   \textbf{0.008} &       \textbf{0.398} &   \textbf{0.005} &     \textbf{ 7.011} &   \textbf{0.000} \\
            SEP     & TL WS Stop Grad.            &      \textbf{3.394} &   \textbf{0.000} &       \textbf{0.091} &   \textbf{0.000} &     \textbf{ 3.640} &   \textbf{0.000} \\
                    & TL WS Full Grad.            &      \textbf{2.040} &   \textbf{0.029} &       \textbf{0.127} &   \textbf{0.000} &     \textbf{13.041} &   \textbf{0.000} \\
                    & TL Stop Grad.     &      \textbf{2.022} &   \textbf{0.031} &       \textbf{0.186} &   \textbf{0.000} &     \textbf{ 2.009} &   \textbf{0.032} \\
                    & TL Full Grad.     &              1.804  &           0.069  &       \textbf{0.217} &   \textbf{0.000} &     \textbf{ 8.013} &   \textbf{0.000} \\
            R$^2$   & TL WSStop Grad.            &      \textbf{4.567} &   \textbf{0.000} &       \textbf{0.514} &   \textbf{0.041} &              1.323  &   		 0.386 \\
                    & TL WSFull Grad.            &      \textbf{2.832} &   \textbf{0.002} &       \textbf{0.404} &   \textbf{0.006} &     \textbf{ 5.199} &   \textbf{0.000} \\
                    & TL Stop Grad.     &      \textbf{3.034} &   \textbf{0.001} &       			1.438 &   0.261  &       		  1.506  &   		 0.205 \\
                    & TL Full Grad.     &      \textbf{2.990} &   \textbf{0.001} &       \textbf{0.268} &   \textbf{0.000} &     \textbf{86.583} &   \textbf{0.000} \\
            BIAS    & TL WS Stop Grad.            &      \textbf{1.946} &   \textbf{0.041} &       \textbf{0.193} &   \textbf{0.000} &     \textbf{ 2.662} &   \textbf{0.003} \\
                    & TL WS Full Grad.            &              1.499  &           0.211  &       			0.851 &   0.617  &              1.488  &   		 0.219 \\
                    & TL Stop Grad.     &      \textbf{5.394} &   \textbf{0.000} &       			0.641 &   0.169  &     \textbf{ 3.216} &   \textbf{0.000} \\
                    & TL Full Grad.     &      \textbf{3.806} &   \textbf{0.000} &       			0.861 &   0.642  &     \textbf{ 6.910} &   \textbf{0.000} \\
            \bottomrule
    \end{tabular}
    \vspace{.1cm}
    \caption{\parbox{.8\linewidth}{Test statistics and p-values for F test on change in variance of performance metrics. Significant tests are highlighted with bold faced numbers. The test is performed such that a test value larger than 1 corresponds to a variance reduction from the baseline to the Weight Share strategy, and vice verca for a test value less than 1.}}
    \label{tab:transfer_F_tests}
\end{table}

The average rankings of the five strategies in terms of the performance metrics RMSE, Sep, $R^2$ and (absolute) bias are shown in Table \ref{tab:transfer_avg_rank}. For all metrics the ranking is 1) Weights Share 2) TL Full Gradient 3) TL WS Full Gradient. Based on these, a Friedman rank test is performed with the test statistics and p-values shown in Table \ref{tab:transfer_friedman_tests}. It is clear that all the tests are significant, meaning that there is a significant grouping for all performance metrics. For the post hoc analysis of the rankings we calculate the critical value for pairwise differences as $CD = 2.728 \cdot \sqrt{5\cdot (5+1)/(6\cdot 40\cdot 3)} = 0.5569$. From this it is clear that for RMSE, SEP and R$^2$, Weight Share is ranked 1, the two Full Gradient strategies are ranked 2 and the two Stop Gradient Strategies takes the last spots. For the Bias, Weight Share and TL Full Gradient are tied for number 1, TL WS Full Gradient and TL WS Stop Gradient are ranked 2 and TL Stop Gradient is ranked 3.

\begin{table}[t]
    \centering
    \begin{subtable}[t]{.6\linewidth}
        \centering
        \begin{tabular}{lrrrr}
            \toprule
            {}  &  RMSE &  SEP &  R$^2$ &  Bias \\
            Strategy    &   &  &  &   \\
            \midrule
            Weight Share        &     \textbf{1.7917} &    \textbf{1.8667} &   \textbf{1.7917} &     \textbf{2.3083} \\
            TL Full Grad.   &     2.3500 &    2.4583 &   2.3500 &     \textbf{2.6583} \\
            TL WS Full Grad.          &     2.6167 &    2.5167 &   2.6167 &     2.8667 \\
            TL Stop Grad.   &     3.9500 &    3.6750 &   3.9500 &     4.0000 \\
            TL WS Stop Grad.          &     4.2917 &    4.4833 &   4.2917 &     3.1667 \\
            \bottomrule
        \end{tabular}
        \caption{\parbox{.8\linewidth}{Average ranking of the 5 strategies evaluated over the three data sets}}
        \label{tab:transfer_avg_rank}
    \end{subtable}
    \hfill
    \begin{subtable}[t]{.3\linewidth}
        \centering
        \begin{tabular}{lrr}
            \toprule
            {} &     Test value & p-value \\
            \midrule
            RMSE & \textbf{101.387} &   \textbf{0.000} \\
            SEP  & \textbf{ 96.087} &   \textbf{0.000} \\
            R2   & \textbf{101.387} &   \textbf{0.000} \\
            BIAS & \textbf{ 23.356} &   \textbf{0.000} \\
            \bottomrule
        \end{tabular}
        \caption{Test statistic and p-value}
        \label{tab:transfer_friedman_tests}
    \end{subtable}
    \caption{Average ranking, test- and p-values from the Friedman rank test. Significant results are highlighted in bold.}
    \label{tab:transfer_friedman}
\end{table}

\section{Conclusion} \label{sec:conclusion}
We have proposed a novel method for training deep convolutional neural networks that learn from multiple data sets containing different numbers of variables using weight sharing. We demonstrated this in two experiments. In the first experiment we combined two medium sized data sets and compared the performance to that of neural nets trained individually on each data set. In the second experiment we combined a medium sized and a small data set, and compared the performance to that of transfer learning from a pre-trained network.

We have showed that when combining two medium sized data sets, this reduces the variance of the produced networks of most of our performance metrics. Furthermore, the proposed strategy produced a significantly smaller prediction error on test samples with the same distribution as the validation set, while for test samples with a different distribution than the validation set (a small shift in wavelengths), the individual trained net performed better.

The proposed method enables training of deep convolutional neural nets though only having few training samples available by co-training with a medium sized data set. Furthermore, it also enables transfer learning without resizing the smaller data set. We showed, that for a small number of training samples, the proposed co-training procedure outperformed both types of transfer learning strategy.

\section*{Acknowledgements}
The research is partially funded by BIOPRO (www.biopro.nu) which is financed by the European Regional Development Fund (ERDF), Region Zealand (Denmark) and BIOPRO partners. We would like to acknowledge the Walloon Agricultural Research Centre (CRA-W, Chaussée de Namur 15, 5030 Gembloux, Belgium) to provide the test data for the Chimiometrie 2018 and 2019 data sets used in the study.

\newpage
\appendix
\section{Architecture} \label{app:Arch}
\begin{table}[H]
    \centering
    \begin{tabular}{lll}
        \toprule
        Layer & Parameters of architecture 1 & Parameters of architecture 2 \\ 
        \midrule
        Input & $\tilde{x}$ & $\tilde{x}$ \\
        Convolution & 8 filters, $1\times 11 strides = 1$ & 8 filters, $1\times 11 strides = 1$ \\
        Max-pooling & $pool\_size=2, strides=2$ & $pool\_size=2, strides=2$  \\
        Batch Normalization & - & - \\
        Convolution & 8 filters, $1\times 11 strides = 1$ & 8 filters, $1\times 11 strides = 1$  \\
        Max-pooling & $pool\_size=2, strides=2$ & $pool\_size=2, strides=2$\\
        Dropout & 1D Spatial, $p_{keep} = 0.95$  & 1D Spatial, $p_{keep} = 0.95$  \\
        Batch Normalization & - & -  \\
        Convolution & 16 filters, $1\times 11 strides = 1$ & 16 filters, $1\times 8 strides = 1$  \\
        Max-pooling & $pool\_size=2, strides=2$ & $pool\_size=2, strides=2$  \\
        Batch Normalization & - & - \\
        Convolution & 16 filters, $1\times 11 strides = 1$ & 16 filters, $1\times 8 strides = 1$  \\
        Max-pooling & $pool\_size=2, strides=2$ & $pool\_size=2, strides=2$  \\
        Dropout & 1D Spatial, $p_{keep} = 0.95$ & 1D Spatial, $p_{keep} = 0.95$  \\
        Batch Normalization & - & - \\
        Convolution & 24 filters, $1\times 6, strides = 1$ & 24 filters, $1\times 6, strides = 1$  \\
        Max-pooling & $pool\_size=2, strides=2$ & $pool\_size=2, strides=2$  \\
        Batch Normalization & - & - \\
        Convolution & 24 filters, $1\times 6, strides = 1$ & 24 filters, $1\times 6, strides = 1$  \\
        Max-pooling & $pool\_size=2, strides=2$ & $pool\_size=2, strides=2$  \\
        Dropout & 1D Spatial, $p_{keep} = 0.95$ & 1D Spatial, $p_{keep} = 0.95$  \\
        Flatten & - & - \\
        Batch Normalization & - & - \\
        \bottomrule
    \end{tabular}
    \vspace{.1cm}
    \caption{Shared CNN architectures.}
    \label{tab:Common_arch_1}
\end{table}

\begin{table}[H]
    \centering
    \begin{tabular}{lrr}
        \toprule
        Data set & FC1 & FC2 \\
        \midrule
        Chimiometrie 2018 & 10 units & 1 unit \\
        Chimiometrie 2019 & 30 units & 3 units \\
        Small data sets & 10 units & 1 unit \\
         \bottomrule
    \end{tabular}
    \vspace{.1cm}
    \caption{Number of hidden units in each of the fully connected layers.}
    \label{tab:fc_part}
\end{table}

\newpage
\section{Additional Results} \label{app:add_res}
\subsection{Experiment 1}
\begin{table}[H]
    \centering
    \begin{tabular}{lcccccc}
        \toprule
        {} & \multicolumn{2}{c}{MAD 2018} & \multicolumn{2}{c}{RMSE 2018} & \multicolumn{2}{c}{WRMSE 2019} \\
        {} & Baseline & Weight Share & Baseline & Weight Share & Baseline & Weight Share \\
        \midrule
        mean &  0.433 & \textbf{0.426} &    0.789 & \textbf{0.771} &    \textbf{0.539} &    0.583 \\
        std  &  0.021 & \textbf{0.017} &    0.039 & \textbf{0.032} &    0.037 & \textbf{0.036} \\
        min  &  \textbf{0.380} &    0.393 & 0.714 & \textbf{0.703} &    \textbf{0.459} &    0.517 \\
        25\%  & 0.421 & \textbf{0.415} & 0.760 & \textbf{0.747} & \textbf{0.512} & 0.552 \\
        50\%  & 0.432 & \textbf{0.425} & 0.782 & \textbf{0.767} & \textbf{0.536} & 0.576 \\
        75\%  & 0.453 & \textbf{0.436} & 0.814 & \textbf{0.786} & \textbf{0.569} & 0.610 \\
        max  &  0.464 & \textbf{0.458} & 0.880 & \textbf{0.863} & \textbf{0.632} & 0.672 \\
        \\
        {} & \multicolumn{2}{c}{Bias 2019 - 1} & \multicolumn{2}{c}{Bias 2019 - 2} & \multicolumn{2}{c}{Bias 2019 -3} \\
        {} & Baseline & Weight Share & Baseline & Weight Share & Baseline & Weight Share \\
        \midrule
        mean &  \textbf{0.089} &    0.118 & \textbf{-0.169} &    -0.220 &    \textbf{-0.165} &    -0.555 \\
        std  &  0.029 & \textbf{0.024} & 0.284 & \textbf{0.222} & \textbf{0.676} & 0.792 \\
        min  &  \textbf{0.004} & 0.058 & -0.708 &    \textbf{-0.624} &    \textbf{-1.814} &    -2.185 \\
        25\%  & \textbf{0.074} & 0.099 & \textbf{-0.330} &    -0.387 &    \textbf{-0.446} &    -0.893 \\
        50\%  & \textbf{0.094} & 0.121 & \textbf{-0.140} &    -0.213 &    \textbf{-0.110} &    -0.505 \\
        75\%  & \textbf{0.107} & 0.134 & \textbf{0.005} & -0.120 &    0.303 & \textbf{-0.015} \\
        max  &  \textbf{0.149} & 0.163 & \textbf{0.298} & 0.373 & \textbf{1.150} & 1.281 \\
        \bottomrule
    \end{tabular}
    \vspace{.1cm}
    \caption{Summary statistics for weight sharing. Bold faced numbers mark the best performance of the two strategies.}
    \label{tab:weight_share_stats}
\end{table}
\newpage

\subsection{Experiment 2}
\begin{table}[H]
    \centering
    \begin{tabular}{lccccc}
        \toprule
        {} & \multicolumn{5}{c}{RMSE}\\
        {} & Weight Share & TL WS Full Gradient & TL WS Stop Gradient & TL Full Gradient & TL Stop Gradient \\
        \midrule
        mean &  \textbf{3.923} & 4.448 & 4.635 & 4.417 & 4.356 \\
        std  &  \textbf{0.218} & 0.324 & 0.401 & 0.337 & 0.336 \\
        min  &  \textbf{3.602} & 3.860 & 3.973 & 3.824 & 3.864 \\
        25\%  & \textbf{3.778} & 4.215 & 4.270 & 4.125 & 4.134 \\
        50\%  & \textbf{3.869} & 4.436 & 4.564 & 4.428 & 4.287 \\
        75\%  & \textbf{4.009} & 4.633 & 4.950 & 4.624 & 4.558 \\
        max  &  \textbf{4.503} & 5.395 & 5.365 & 5.213 & 5.192 \\
        \\
        {} & \multicolumn{5}{c}{SEP}\\
        {} & Weight Share & TL WS Full Gradient & TL WS Stop Gradient & TL Full Gradient & TL Stop Gradient \\
        \midrule
        mean &  \textbf{3.904} & 4.391 &   4.554 & 4.329 &    4.161 \\
        std  &  \textbf{0.219} & 0.313 &   0.403 & 0.294 &    0.311 \\
        min  &  \textbf{3.568} & 3.858 &   3.897 & 3.754 &    3.748 \\
        25\%  & \textbf{3.753} & 4.160 &   4.224 & 4.071 &    3.958 \\
        50\%  & \textbf{3.854} & 4.379 &   4.499 & 4.313 &    4.057 \\
        75\%  & \textbf{3.996} & 4.575 &   4.877 & 4.560 &    4.206 \\
        max  &  \textbf{4.501} & 5.243 &   5.364 & 4.940 &    5.186 \\
        \\
        {} & \multicolumn{5}{c}{R$^2$}\\
        {} & Weight Share & TL WS Full Gradient & TL WS Stop Gradient & TL Full Gradient & TL Stop Gradient \\
        \midrule
        mean &  \textbf{0.938} & 0.920 &    0.913 &   0.921 &   0.923 \\
        std  &  \textbf{0.007} & 0.012 &    0.015 &   0.012 &   0.012 \\
        min  &  \textbf{0.918} & 0.883 &    0.884 &   0.890 &   0.891 \\
        25\%  & \textbf{0.935} & 0.913 &    0.901 &   0.914 &   0.916 \\
        50\%  & \textbf{0.940} & 0.921 &    0.916 &   0.921 &   0.926 \\
        75\%  & \textbf{0.942} & 0.928 &    0.926 &   0.931 &   0.931 \\
        max  &  \textbf{0.948} & 0.940 &    0.936 &   0.941 &   0.940 \\
        \\
        {} & \multicolumn{5}{c}{Bias}\\
        {} & Weight Share & TL WS Full Gradient & TL WS Stop Gradient & TL Full Gradient & TL Stop Gradient \\
        \midrule
        mean &  \textbf{0.104} & 0.558 & 0.690 & 0.519 & 0.961 \\
        std  &  \textbf{0.377} & 0.462 & 0.526 & 0.736 & 0.876 \\
        min  &  -0.877 & \textbf{-0.155} &   -0.739 &    -1.049 &    -0.646 \\
        25\%  & -0.077 &    0.220 & 0.350 & \textbf{0.038} & 0.315 \\
        50\%  & \textbf{0.128} &    0.539 &  0.707 & 0.370 & 1.044 \\
        75\%  & \textbf{0.339} &    0.743 &  1.039 & 1.073 & 1.360 \\
        max  &  \textbf{0.740} & 1.770 & 1.801 & 2.573 & 3.365 \\
        \bottomrule
    \end{tabular}
    \vspace{.1cm}
    \caption{Summary statistics for transfer learning for the IDRC 2002 data set. Bold faced numbers are the per line best performance.}
    \label{tab:idrc_transfer_stats}
\end{table}
\newpage

\begin{minipage}[!t]{\linewidth} 
\centering
\begin{table}[H]
    \centering
    \begin{tabular}{lrrrrrr}
        \toprule
        {} & \multicolumn{5}{c}{RMSE}\\
        {} & Weight Share & TL WS Full Gradient & TL WS Stop Gradient & TL Full Gradient & TL Stop Gradient \\
        \midrule
        mean &  0.632 &    0.538 &    1.303 &    \textbf{0.495} &    1.292 \\
        std  &  0.097 &    0.071 &    \textbf{0.033} &    0.061 &    0.058 \\
        min  &  0.454 &    \textbf{0.390} &    1.267 &    0.399 &    1.179 \\
        25\%  & 0.569 &   0.487 &   1.283 &   \textbf{0.459} &   1.249 \\
        50\%  & 0.625 &   0.519 &   1.295 &   \textbf{0.492} &   1.289 \\
        75\%  & 0.688 &   0.582 &   1.313 &   \textbf{0.517} &   1.327 \\
        max  &  0.830 &    0.732 &    1.453 &    \textbf{0.692} &    1.445 \\
        \\
        {} & \multicolumn{5}{c}{SEP}\\
        {} & Weight Share & TL WS Full Gradient & TL WS Stop Gradient & TL Full Gradient & TL Stop Gradient \\
        \midrule
        mean &  0.595 & 0.480 & 1.297 & \textbf{0.466} & 1.140 \\
        std  &  0.093 & 0.033 & \textbf{0.028} & 0.043 & 0.040 \\
        min  &  0.428 & 0.390 & 1.264 & \textbf{0.372} & 1.071 \\
        25\%  & 0.535 & 0.462 & 1.280 & \textbf{0.440} & 1.115 \\
        50\%  & 0.597 & 0.477 & 1.292 & \textbf{0.463} & 1.133 \\
        75\%  & 0.655 & 0.501 & 1.307 & \textbf{0.493} & 1.160 \\
        max  &  0.805 & \textbf{0.562} & 1.409 & 0.569 & 1.246 \\
        \\
        {} & \multicolumn{5}{c}{R$^2$}\\
        {} & Weight Share & TL WS Full Gradient & TL WS Stop Gradient & TL Full Gradient & TL Stop Gradient \\
        \midrule
        mean &  0.865 & 0.903 & 0.440 & \textbf{0.918} &    0.448 \\
        std  &  0.041 & 0.026 & 0.029 & \textbf{0.021} &    0.049 \\
        min  &  0.773 & 0.823 & 0.304 & \textbf{0.842} &    0.311 \\
        25\%  & 0.844 & 0.888 & 0.431 & \textbf{0.912} &    0.419 \\
        50\%  & 0.871 & 0.911 & 0.446 & \textbf{0.920} &    0.451 \\
        75\%  & 0.893 & 0.922 & 0.457 & \textbf{0.930 }&    0.485 \\
        max  &  0.932 & \textbf{0.950} & 0.470 & 0.948 &    0.541 \\
        \\
        {} & \multicolumn{5}{c}{Bias}\\
        {} & Weight Share & TL WS Full Gradient & TL WS Stop Gradient & TL Full Gradient & TL Stop Gradient \\
        \midrule
        mean &  0.133 &    -0.196 & -0.095 & \textbf{0.069} & -0.593 \\
        std  &  0.172 &    0.159 &  \textbf{0.076} &  0.160 &  0.138 \\
        min  &  -0.349 &    -0.521 &    -0.354 &    \textbf{-0.258} &    -0.864 \\
        25\%  & 0.049 &   -0.319 &  -0.139 &  \textbf{-0.013} &  -0.683 \\
        50\%  & 0.127 &   -0.193 &  -0.103 &  \textbf{0.062} &  -0.619 \\
        75\%  & 0.207 &   -0.102 &  \textbf{-0.054} &  0.164 &  -0.530 \\
        max  &  0.565 &    0.208 &  \textbf{0.072} &  0.451 &  -0.287 \\
        \bottomrule
    \end{tabular}
    \vspace{.1cm}
    \caption{Summary statistics for transfer learning for the Wheat data set. Bold faced numbers are the per line best performance.}
    \label{tab:wheat_transfer_stats}
\end{table}
\end{minipage}

\vfill

\begin{table}[H]
    \centering
    \begin{tabular}{lccccc}
        \toprule
        {} & \multicolumn{5}{c}{RMSE}\\
        {} & Weight Share & TL WS Full Gradient & TL WS Stop Gradient & TL Full Gradient & TL Stop Gradient \\
        \midrule
        mean & \textbf{0.635} &  0.855 & 1.480 & 2.187 & 1.442 \\
        std  & \textbf{0.245} &  0.821 & 0.428 & 0.648 & 0.351 \\
        min  & \textbf{0.271} &  0.363 & 0.832 & 1.115 & 0.837 \\
        25\%  & \textbf{0.459} & 0.592 & 1.196 & 1.870 & 1.192 \\
        50\%  & \textbf{0.621} & 0.703 & 1.484 & 2.107 & 1.388 \\
        75\%  & \textbf{0.749} & 0.830 & 1.614 & 2.423 & 1.640 \\
        max  & \textbf{1.600} &  5.635 & 3.492 & 3.932 & 2.620 \\
        \\
        {} & \multicolumn{5}{c}{SEP}\\
        {} & Weight Share & TL WS Full Gradient & TL WS Stop Gradient & TL Full Gradient & TL Stop Gradient \\
        \midrule
        mean & \textbf{0.577} &  0.798 & 1.298 & 2.067 & 1.312 \\
        std  & \textbf{0.223} &  0.807 & 0.426 & 0.632 & 0.317 \\
        min  & \textbf{0.264} &  0.283 & 0.549 & 0.655 & 0.752 \\
        25\%  & \textbf{0.416} & 0.553 & 1.058 & 1.767 & 1.128 \\
        50\%  & \textbf{0.575} & 0.635 & 1.208 & 2.017 & 1.284 \\
        75\%  & \textbf{0.677} & 0.788 & 1.491 & 2.286 & 1.584 \\
        max  & \textbf{1.563} &  5.464 & 3.333 & 3.826 & 2.228 \\
        \\
        {} & \multicolumn{5}{c}{R$^2$}\\
        {} & Weight Share & TL WS Full Gradient & TL WS Stop Gradient & TL Full Gradient & TL Stop Gradient \\
        \midrule
        mean & \textbf{0.981} &  0.961 & 0.927 & 0.429 &     0.928 \\
        std  & \textbf{0.045} &  0.102 & 0.051 & 0.414 &     0.055 \\
        min  & 0.713 &  0.484 & \textbf{0.758} & -1.311 &    0.745 \\
        25\%  & \textbf{0.983} & 0.981 & 0.907 & 0.312 &     0.903 \\
        50\%  & \textbf{0.989} & 0.986 & 0.936 & 0.515 &     0.944 \\
        75\%  &\textbf{ 0.994} & 0.991 & 0.965 & 0.684 &     0.966 \\
        max  & \textbf{0.998} &  0.997 & 0.983 & 0.935 &     0.985 \\
        \\
        {} & \multicolumn{5}{c}{Bias}\\
        {} & Weight Share & TL WS Full Gradient & TL WS Stop Gradient & TL Full Gradient & TL Stop Gradient \\
        \midrule
        mean &  0.051 &     \textbf{0.048} &     -0.549 &    0.049 &     -0.366 \\
        std  &  \textbf{0.280} &     0.342 &     0.457 &     0.736 &     0.502 \\
        min  & \textbf{-0.724} &     -1.379 &    -1.354 &    -1.436 &    -1.377 \\
        25\%  & \textbf{-0.089} &    -0.095 &    -0.838 &    -0.566 &    -0.710 \\
        50\%  &  \textbf{0.015} &    0.066 &     -0.558 &    0.068 &     -0.277 \\
        75\%  &  0.247 &    0.230 &     -0.209 &    0.601 &     \textbf{-0.027} \\
        max  &  0.598 &     0.697 &     \textbf{0.547} &     1.578 &     0.731 \\
        \bottomrule
    \end{tabular}
    \vspace{.1cm}
    \caption{Summary statistics for transfer learning for the SWRI data set. Bold faced numbers are the per line best performance.}
    \label{tab:swri_transfer_stats}
\end{table}
\newpage

\sectionmark{Bibliography}
\addcontentsline{toc}{section}{Bibliography}        
\bibliography{references.bib}

\end{document}